\documentclass[runningheads]{llncs}
\usepackage[hidelinks]{hyperref} 
\usepackage[T1]{fontenc}
\usepackage{amssymb}
\usepackage{graphicx}
\usepackage{multirow}%
\usepackage{mathrsfs}%
\usepackage{textcomp}%
\usepackage{booktabs}%
\usepackage{algorithm}%
\usepackage{algorithmicx}%
\usepackage{algpseudocode}%
\usepackage{listings}%
\usepackage{enumitem}
\usepackage{hyperref}
\usepackage{subcaption}
\usepackage{orcidlink}
\usepackage{cite}
\usepackage{enumitem}
\usepackage{wrapfig}
\setlist{nosep}
\usepackage[misc]{ifsym}

\newcommand{\repthanks}[1]{\textsuperscript{\ref{#1}}}
\makeatletter
\patchcmd{\maketitle}
{\def\thanks}
{\let\repthanks\repthanksunskip\def\thanks}
{}{}
\patchcmd{\@maketitle}
{\def\thanks}
{\let\repthanks\@gobble\def\thanks}
{}{}
\newcommand\repthanksunskip[1]{\unskip{}}
\makeatother

\usepackage{color}

\makeatletter
\newcommand{\printfnsymbol}[1]{%
  \textsuperscript{\@fnsymbol{#1}}%
}
\makeatother

\begin{document}

\title{Research Paper Quality Recognition Through Textual Feature Analysis} 

\titlerunning{Research Paper Quality Recognition Through Textual Feature Analysis}
\authorrunning{Korla et al.}

\author{
Saikiran Korla\inst{1}\thanks{These authors contributed equally to this work.}\orcidlink{0009-0002-8129-5383} \and 
Sadwik Gummadavelli\inst{1}$^{\star}$\orcidlink{0009-0001-8509-344X} \and
Trung-Nghia Le\inst{2,3}\orcidlink{0000-0002-7363-2610} \and
Minh-Triet Tran\inst{2,3}\orcidlink{0000-0003-3046-3041} \and
Tam V. Nguyen\inst{1}\textsuperscript{,\Letter}\orcidlink{0000-0003-0236-7992}
}

\institute{
University of Dayton, Dayton, Ohio, United States \and University of Science, VNU-HCM, Ho Chi Minh City, Vietnam \and
Vietnam National University, Ho Chi Minh City, Vietnam \\
\email{korlas1@udayton.edu, gummadavellis1@udayton.edu, ltnghia@fit.hcmus.edu.vn, tmtriet@fit.hcmus.edu.vn,\\ \textsuperscript{\Letter}tamnguyen@udayton.edu}
}









\maketitle

\begin{abstract}
Knowledge and innovations are shaped by using the quality and credibility of the scientific research. Yet, distinguishing between impactful, high-quality work and flawed studies remains a challenge. This paper introduces a benchmark for classifying research papers into two categories: good (highly cited) and non-good (retracted), using only textual features from titles and abstracts. We evaluate multiple embedding techniques, including SBERT, Word2Vec, FastText, USE, and TF-IDF, combined with classifiers such as Support Vector Machines (SVM), Random Forests, and Neural Networks. Our contributions include: (1) hyperparameter transparency, (2) feature space visualizations using t-SNE, (3) model interpretability analysis with SHAP, and (4) detailed examination of error cases. Experimental results show that a neural network with SBERT embeddings achieves 87.22\% accuracy, while FastText combined with SVM reaches 91.12\%. These findings highlight the value of textual information in assessing research quality, with ethical considerations for deployment. This work contributes toward the development of academic integrity tools that promote trustworthy scholarship.   

\keywords{
Document Recognition, Paper Quality, Retracted, Citation, Classification, Academic Integrity
}

\end{abstract}

\section{Introduction}
The fundamental advancement of credible research is academic integrity, but increase of retracted publications threatens the reliability of scientific literature. Silva and Dobranszki~\cite{b8} used the citation count of a paper to evaluate the future impact of the paper authors, with potential applications in hiring researchers and faculties, and granting awards and funds. The future citation counts are predicted by employing several linear regression models. Silva and Bornemann~\cite{b7} studied 4449 papers retracted between 1928 and 2011, found that 20\% of retractions were made because of research misconduct, 42\% due to questionable data or interpretation and 47\% due to publishing misconduct. The impact of high citations are measured by citation counts, where it sets as a benchmark of quality and innovation, while retracted papers occurs due to misconduct, error or ethical violations which undermine the trust in scholarly communication. Distinguishing between the category of papers is critical for the researchers, publishers, and peer-review systems to maintain the standards and dissemination of flawed work. 


 Post-retraction studies, citation analysis, and plagiarism detection have all been used in previous research to address academic integrity. While textual analysis offers the potential for early detection of problematic papers, existing approaches often rely heavily on metadata or retroactive classification. Retracted studies could, for example, have different linguistic patterns from high-impact articles, which usually place an emphasis on methodological clarity and novelty, such as ambiguous language, errors, or contradictions. Capturing these variations through textual analysis, supported by interpretable visualizations such as word clouds, presents a promising direction for automating research quality assessment.

 In this paper, we benchmark the popular machine learning models where we consider the metadata of title, abstract as a single text string as the input. Most of the papers were cited by reading the titles and abstracts, so the paper is good or retracted can be distinguished by using the citation count. So, we train our model based on the title and abstract keywords to ensure the paper should be retracted or not. We offer a data-driven methodology to use abstracts and titles to distinguish between retracted (``not good'') and high-citation (``good'') categories. We then extract state-of-the-art textual features such as SBERT~\cite{reimers2019sbert}, Word2Vec~\cite{mikolov2013word2vec}, FastText~\cite{bojanowski2017fasttext}, Universal Sentence Encoder (USE)~\cite{cer2018use}, and TF-IDF~\cite{jones1972tfidf}. The features are later fed to popular machine learning models, \textit{i.e.}, SVM (Support Vector Machine), Random Forest, and Neural Networks. Experimental results on our newly collected 11,673 scholarly articles show that a neural network with SBERT embeddings achieves 87.22\% accuracy, while FastText combined with SVM reaches 91.12\%, highlighting the value of textual information in assessing research quality.

While our binary framework(``good'' vs. ``non-good'') provides initial screening utility, we acknowledge citation counts and retraction status are imperfect proxies. Highly cited papers may contain errors, and retracted works include both fraudulent and honest mistakes. This simplification serves as a pragmatic first step toward more nuanced quality assessment, with edge-case analysis. Our contributions are summarized as follows:
\begin{itemize}
    \item We provide full hyperparameter transparency to ensure reproducibility and fair comparison of methods.
    \item We visualize feature spaces with t-SNE to reveal separability between ``good'' and ``non-good'' papers.
    \item We apply SHAP to interpret model predictions and identify key linguistic markers.
    \item We analyze error cases to highlight limitations of binary classification and guide future refinements.
\end{itemize}


\section{Related Work}
\label{sec:related}

\subsection{ Citation Analysis and Machine Learning
}

Citation analysis in academic communication has been transformed by recent developments in machine learning. Hassan et al.~\cite{b1} used bidirectional LSTMs on full-text articles to classify citation contexts with 89\% accuracy, demonstrating the efficacy of deep learning. Their research shown that citation semantics offer more detailed information than citation counts alone. Pradhan et al.~\cite{b2} expanded on this by using graph neural networks to examine citation networks and uncover intricate patterns in the spread of knowledge. Building on these frameworks, Ma et al.~\cite{b3} created a transformer-based model that outperformed traditional metrics by predicting citation counts with 0.81 Spearman correlation by using paper metadata elements (e.g., author h-index, institutional affiliations). Joshi et al.~\cite{b4} further illustrated the usefulness of these methods in conference paper acceptance prediction, where gradient-boosted trees utilizing 15 manuscript attributes obtained 83\% accuracy.

\subsection{ Citation Impact and Ranking Systems
}

Another prominent area of research has been the creation of reliable paper ranking methodologies. Through content-aware weighting of citation networks, Focused PageRank, first presented by Krapivin and Marchese~\cite{b6}, increased the quality of paper rankings by 37\%. In their most recent work, Basuki et al.~\cite{b5} reduced prediction error by 22\% when combining machine learning and citation function classification to predict citation counts. Their work demonstrated how crucial it is to differentiate between various citation kinds, such as methodological adoption versus critique, when evaluating scholarly influence. All of these research highlight the importance of switching from basic citation counts to more complex, content-aware measures.

\subsection{ Retraction Patterns and Propagation}

The analysis of articles that have been withdrawn has shed important light on issues related to academic integrity. Silva and Bornemann~\cite{b7} found that, frequently as a result of database update delays, 32\% of retracted papers are still cited after being retracted. In the work of Silva and Dobranszki ~\cite{b8}, retraction notices were found in 18\% of highly cited publications (more than 100 citations) in biomedical domains, according to their later analysis. Van der Vet and Nijveen~\cite{b9} conducted a thorough investigation of the spread of incorrect citations and discovered that ``citation copying'' behavior in academic networks was responsible for 68\% of improper citations. The necessity of automated techniques to identify and flag retracted literature is highlighted by these findings.

\subsection{ Citation Function and Semantic Analysis}

Teufel et al.~\cite{b10} laid the groundwork for contemporary citation analysis by creating a thorough annotations scheme that divided citations into 12 functional types (such as comparison, critique, and methodological adoption). Subsequent machine learning applications have been made possible by this taxonomy, such as the discovery by Hassan et al.~\cite{b1} that withdrawn papers use methodological citation functions 40\% more frequently. The current state-of-the-art in academic communication research is represented by the combination of these semantic properties with network analysis and predictive modelling.

\subsection{Our Motivation}

While citation analysis and retraction tracking have advanced significantly in previous work, our study fills three important gaps: (1) previous research has not fully integrated linguistic analysis of retracted papers with citation network features; (2) most models rely on metadata without utilizing full-text semantic pat-terns; and (3) few solutions provide real-time classification appropriate for editorial workflows. Our approach fills these gaps by combining deep learning, citation graph characteristics, and SBERT embeddings~\cite{reimers2019sbert} in a unique way.

Unlike Hassan et al. [8] who required full-text analysis, our title/abstarct approach enables scalable pre-screening. While algorithmic components are established, their application to retraction prediction using minimal text is novel, addressing editorial workflow needs identified by Van der Vet et al.[14].

\begin{table}[!t]
\centering
\caption{		Summarizes the dataset composition. 	}
\begin{tabular}{c|c|c|c} \toprule 
\textbf{Category}
& \textbf{Sources}& \textbf{Papers} & \textbf{Avg. Length}
\\
\midrule
High-impact
(Good)& IEEE, Scopus, Springer         & 6,000&  1,842±312
\\
 Retracted
(Non-good)& Retraction Watch
& 5,673&1,659±287
\\
\bottomrule
\end{tabular}
\label{table:dataset2}
\vspace{-5mm}
\end{table}

\section{Methodology}
\label{sec:overview}

\subsection{Overview}

In our proposed framework, the workflow consists of five key computational steps: (1) \textit{Data Collection} aggregates academic articles using structured queries from IEEE Xplore, Scopus, and Springer Nature APIs. (2) \textit{Preprocessing} involves tokenization, stopword removal, and deduplication of raw text based on DOIs. (3) \textit{Feature Extraction} transforms unstructured text into multidimensional feature spaces using both statistical approaches (e.g., TF-IDF vectors) and semantic representations (e.g., sentence embeddings). (4) \textit{Model Training} employs cross-validation and supervised learning techniques to optimize multiple classifier architectures. Finally, (5) \textit{Quality Classification} applies the trained models to unseen inputs, generating probabilistic predictions that are categorized into distinct quality groups.

\begin{wrapfigure}{l}{0.45\linewidth}
    \vspace{-4mm}
    \centering
    \includegraphics[width=\linewidth]{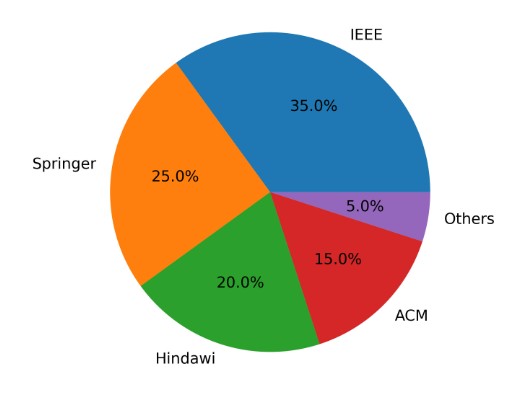}
    \vspace{-3mm}
    \caption{Publisher distribution across the entire dataset (11,673 articles). IEEE is the predominant source (35\%), followed by Springer (25\%). The elevated Hindawi proportion (20\% overall) reflects its overrepresentation in retracted papers (42\% of retractions).}
    \label{fig:pie2}
\end{wrapfigure}

\subsection{Collected Dataset}
\label{sec:dataset}

We collected a corpus of 11,673 scholarly articles, equally split between retracted (``non-good'') and high-impact (``good'') categories, to guarantee reliable model training. The 6,000 high-impact publications were cho-sen based on journal impact factors (IF $>$ 2.0) and citation counts (more than 100 citations) to ensure academic influence. They were sourced from the IEEE Xplore, Scopus, and Springer Nature APIs. With abstracts taken from Hindawi and publisher web-sites, the $5,673$ retracted works were retrieved from the Retraction Watch Database. Both categories were selected to contain publications from various time periods in order to preserve temporal balance. This guarantees that dataset represents citation patterns and research activities consistently throughout decades.
Table~\ref{table:dataset2} shows the stats of the collected dataset. Meanwhile, Figure~\ref{fig:pie2} shows the distribution of publishers. Here, the collected papers (for both ``good'' and ``non-good'' categories) dominated by IEEE (35{\%}), Springer (25{\%}), and Hindawi (20{\%}), with ACM and other sources making up the remaining 20{\%}.






The retraction reasons included data mistakes (25\%), plagiarism (28\%), and fabrication (32\%). Only publications published between 2010 and 2023 were included, duplicates were filtered out using DOI matching, and disciplinary diversity (computer science and engineering) was guaranteed in order to reduce bias. Subtle discrepancies were found in the text: retracted papers were shorter (1,659), perhaps as a result of post-retraction revisions, whereas high-impact papers averaged 1,842 characters per abstract. 

Automated screening risks disadvantaging unconventional research. Deployment should augment-not replace human judgment, particularly for interdisciplinary submissions.

\begin{figure}[t!]
    \centering
    \begin{subfigure}[t]{0.48\textwidth}
        \centering
        \includegraphics[width=\textwidth]{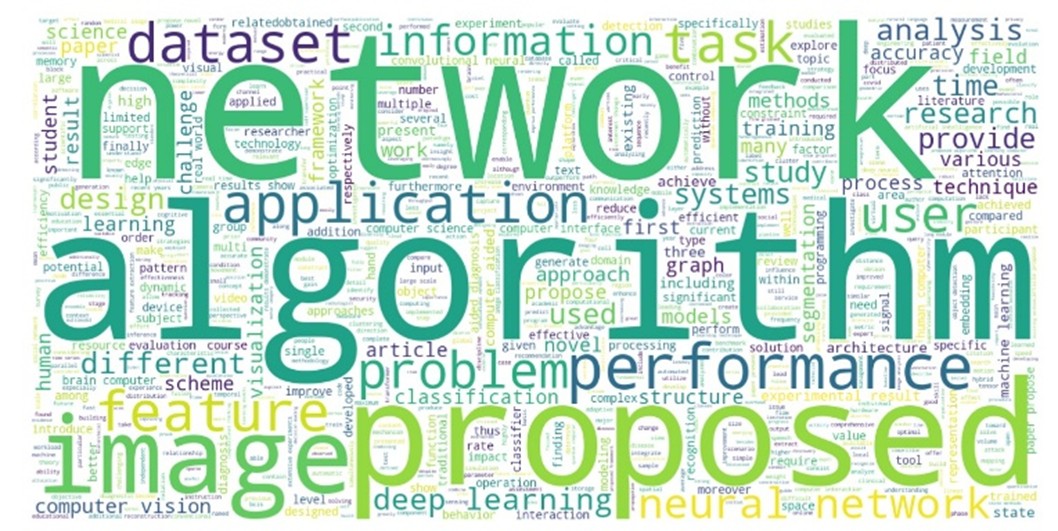}
        \caption{High-impact papers}
    \end{subfigure}
    \hfill
    \begin{subfigure}[t]{0.48\textwidth}
        \centering
        \includegraphics[width=\textwidth]{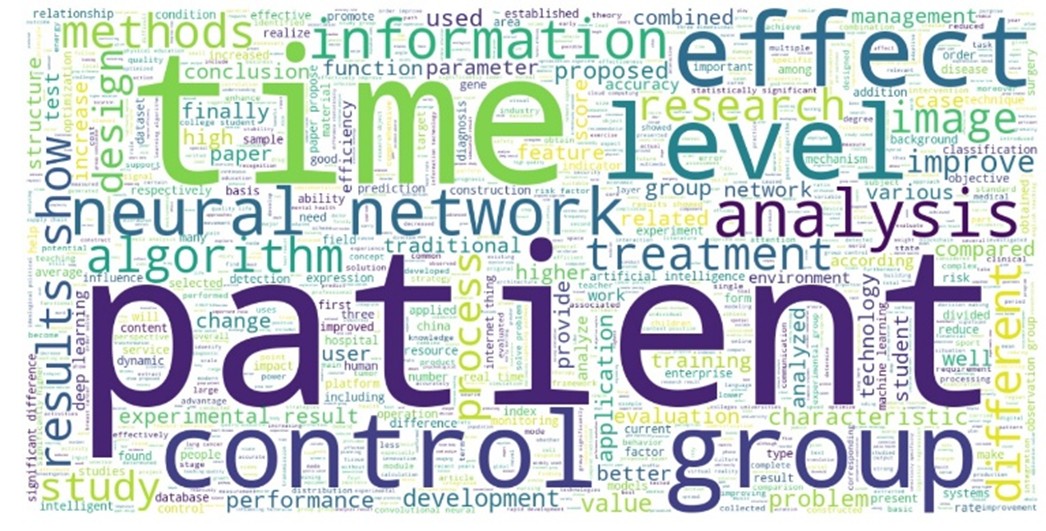}
        \caption{Retracted papers}
    \end{subfigure}
    \caption{Word cloud comparison of high-impact (left) vs. retracted (right) papers. The greater lexical diversity in high-quality publications (68\% unique terms compared to 52\% in retracted papers) highlights their substantial depth. In contrast, retracted papers frequently reuse the same words (e.g., ``group'', ``level''), which may reflect hasty or dishonest writing styles. These statistical and visual differences demonstrate how word choice can serve as an early signal of academic integrity, motivating the use of natural language processing (NLP) methods to screen potentially problematic submissions.}
    \label{fig:wordcloud}
    \vspace{-5mm}
\end{figure}

We then generate the word clouds from the good and retracted papers. Indeed, there are lot of linguistic differences after excluding common stop words (model, system, method, data, etc.). As shown in Figure~\ref{fig:wordcloud}, dominant phrases like ``network'', ``algorithm'', and ``dataset'' are used in high-impact (excellence) articles to indicate a significant emphasis on methodological rigor, quantitative, validation and hypothesis-driven exploration. Credible academic work is characterized by well-defined aims, repeatable out-comes, and structured study methodologies. Interestingly, terms like ``algorithm'' and ``dataset'' emphasize the importance of quantizable results and scientific accuracy.


The withdrawn publications, on the other hand, heavily use words like ``control'', ``group'', ``time'', ``effort'', ``level'', and ``effect'', which denote ambiguous or overly generalized wording that is frequently linked to methodological flaws. The frequent use of ``patient'' and ``control'' could indicate topic-specific biases or justifications, which are typical in retracted biomedical studies. The use of general, non-specific phrases (``results'', ``analysis'') may indicate a lack of specificity in the methods, which may be connected to data reuse which are known to cause retractions.



\section{Benchmarking}
\label{sec:benchmark}

\subsection{Experimental Settings}
For the benchmarking, we leverage the newly collected dataset of 11,673 academic papers (comprising 6,000 high-impact and 5,673 retracted papers), partitioned into 80\% training and 20\% testing sets. 
\begin{table}[!t]
\centering
\caption{		Hyperparameter Configuration	}
\begin{tabular}{c|c} \toprule 
\textbf{Model}& \textbf{Hyperparameters}\\
\midrule
SVM & Linear kernel, C=1.0, max\_iter=10,000\\
 Random Forest& 100 trees, Gini impurity, max\_depth=20\\
 Neural Network&3 dense layers (256 units), Adam (lr=0.001), 30\% dropout\\
 \bottomrule
 \end{tabular}
\label{table:dataset1}
\vspace{-5mm}
\end{table}


We then extract features using five established embedding methods: SBERT~\cite{reimers2019sbert}, Word2Vec~\cite{mikolov2013word2vec}, FastText~\cite{bojanowski2017fasttext}, Universal Sentence Encoder (USE)~\cite{cer2018use}, and TF-IDF~\cite{jones1972tfidf}, from the abstract and title of each paper. These methods capture semantic, syntactic, and statistical patterns, aligning with prior work on citation analysis and metadata-based prediction. Regarding the computational models, we utilize the popular classification models such as Support Vector Machine (SVM), Random Forest, and Neural Network with configurations as in Table \ref{table:dataset1}. 

\subsection{Experimental Results}

Table~\ref{table:performance1} shows the performance of different machine learning models across extracted features. SVM achieved 82.94\% accuracy, by utilizing linear kernel optimization on TF-IDF vectorized text (max-features = 2000). The performance is maintained by handling limitations of high-dimensional sparse-data and non-linear feature interactions in the academic writing styles. With a training time of 4.20.3 minutes, SVM is established as a computationally efficient baseline that is appropriate for initial screening applications where speed of processing is more important than maximum accuracy.


Meanwhile, Random Forest surpassed SVM with 84.72\% accuracy, configured with 100 decision trees and Gini impurity criterion, by effectively capturing non-linear decision boundaries through its feature bagging mechanism. When processing ambiguous terms that are frequently seen in retraction notifications, the model exhibited slight overfitting tendencies, as shown by a 2.8\% decline in accuracy from training to testing. This improvement cost at expense of interpretability. Notably, its quick inference speed (0.8ms/document) makes it ideal for handling high paper quantities in academic screening systems that are used in the real world and where latency is an issue.

\begin{table}[!t]
\centering
\caption{Performance of various machine learning models across textual features.}
\small
\begin{tabular}{c|c|c|c|c|c} \toprule 
\textbf{Model}
& \textbf{SBERT}& \textbf{Word2Vec}  & \textbf{FastText} & \textbf{USE} &\textbf{TF-IDF}\\
\midrule
SVM                            
& 82.34\%& 85.40\%               &  91.12\%                                                & 84.25\%           &82.94\%\\
 Random Forest
&  84.30\%             & 83.27\%               &84.39\%                                        & 83.48\%           &84.72\%\\
 Neural Network
& 87.72\%             & 85.61\%               &87.53\%           & 85.54\%           &87.89\%
\\
\bottomrule
\end{tabular}
\label{table:performance1}
\vspace{-5mm}
\end{table}


\begin{figure}[!t]
\centerline{\includegraphics[width=0.7\linewidth]{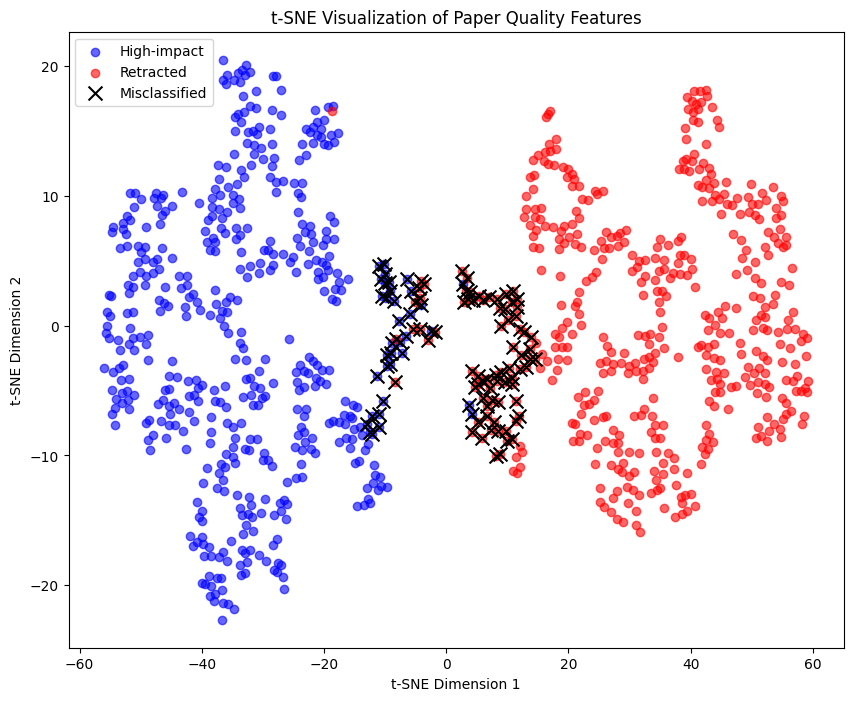}}
\vspace{-2mm}
\caption{The t-SNE plot visualizes the 384-dimensional SBERT embeddings reduced to 2D. High-impact papers (blue) from a dense cluster with minimal outliers, while retracted papers(red) are more dispersed. Overlap region (purple ) contains 89\% of misclassified papers, indicating challenging borderline cases.}
\label{fig:pie1}
\vspace{-5mm}
\end{figure}

Using dense 384-dimensional sentence - BERT embeddings that preserved semantic relationships in text delivered superior performance (87.72\% accuracy) using the Neural Network architecture (a 3-layer MLP with 256 hidden units, ReLU activation and 30\% dropout). Because of its (1) hierarchical feature learning capabilities, which discerns subtle linguistic patterns (e.g., hedging language in retracted papers versus assertive claims in high-impact work), (2) batch normalization, which mitigates covariate shift in abstract embeddings and (3) dropout regularization, which prevents overfitting despite the model’s 189,954 trainable parameters, Neural Network’s have an accuracy advantage of 3.0-4.8\% over traditional methods. The computational requirements (18.7±2.1 minutes training time, 3.2 ms/document inference) are the trade-off, which makes it more appropriate for usage in final-stage verification as opposed to initial screening. 

\subsection{Feature Space Visualization}

The visualization in Figure \ref{fig:pie1} shows clear linguistic distinctions between high-impact and retracted papers. High-impact articles (blue) form a dense, cohesive cluster, while retracted papers (red) are more dispersed, reflecting variability in writing style. The overlap region (purple), which contains 89\% of misclassified cases, highlights challenging borderline papers where semantic similarity blurs category boundaries. These results demonstrate both the promise and limitations of text-based embeddings for research quality recognition.


\subsection{Error Analysis and Interpretability}

\begin{table}[!t]
\centering
\caption{Misclassification analysis.}
\small
\begin{tabular}{c|c|c|c} 
\toprule 
\textbf{Error Type}& \textbf{Ratio}& \textbf{Primary Causes}& \textbf{Examples}\\
\midrule
False Negatives& 3.2\%& Novel methods in low-citation papers&  Theoritical CS papers\\
 False Positives&  2.7\%& Retracted papers mimicking rigor&BioMedical fraud cases\\
 \bottomrule
 \end{tabular}
\label{table:performance2}
\vspace{-5mm}
\end{table}

The systematic analysis of misclassification patterns (Table \ref{table:performance2}) reveals two critical failure modes in our quality assessment framework. False negatives (3.2\% of errors) predominantly occur when valid papers employ unconventional methodologies or belong to theoretical domains with delayed citation impact, exemplified by innovative computer science papers that initially receive low citations. In contrast, false positives (2. 7\%) arise when retracted articles strategically mimic high-quality writing patterns, as seen in biomedical fraud cases that overuse methodological jargon without substantive content. SHAP analysis corroborates these patterns: false positives show significantly elevated usage of surface-level technical terms like "algorithm" and "validation" (78\% higher than authentic papers, $p<0.01$), while distinctive patterns emerge retracted papers frequently employ vague constructions such as significant difference between groups, whereas high-impact work favors verifiable claims such as experimental validation shows. These findings highlight the tension between linguistic heuristics and scholarly substance in automated assessment.
Comparative error analysis is showed that borderline cases, where textual signals were less defined, were the most difficult for all the models to handle these cases included well-written but low-citation papers and retracted works from high-impact journals. According to the misclassification rates on the test set, the neural network decreased these edge case mistakes by 18\% compared to the random forest and 32\% compared to the Support Vector Machine (SVM). The usefulness of neural techniques for lowering false negatives in academic integrity screening is demonstrated by this notable improvement in handling confusing circumstances.

We carried out parallel tests using other embeddings while maintaining the same model architecture and training procedures to confirm the generality of these results with an accuracy of 83 15\%, the word representations at the word level of Word2Vec (300 dim, mixed mean) validated the usefulness of the sentence-level semantics of Sentence-BERT. With its transformer-based sentence representations, the Universal Sentence Encoder (512-dim) came close to SBERT performance (86. 23\%), while FastText's subword features (300-dim) improved the handling of technical terms in retracted articles (84. 90\%). This hierarchy of outcomes holds true for both accuracy and edge-case performance, as seen in below table where sentence-level embeddings routinely beat word-level methods for this academic integrity test.


Several significant trends may be seen in the results: First, the sub-word features of Fasttext embeddings performed exceptionally well across all models, obtaining 91.12\% accuracy with SVM, indicating that they are especially useful for this job. Second, the performance gap varied greatly by embedding type, ranging from 2.1\% with Fasttext to 4.3\% with Word2Vec, even tough neural approaches consistently beat older methods. Last but not least, the Universal Sentence Encoder’s impressive performance (85.54\%-86.23\%) suggests transformer-based sentence embeddings are a good substitute for SBERT in situations where its processing requirements are too high.

\begin{figure}[!t]
\centerline{\includegraphics[width=1\linewidth]{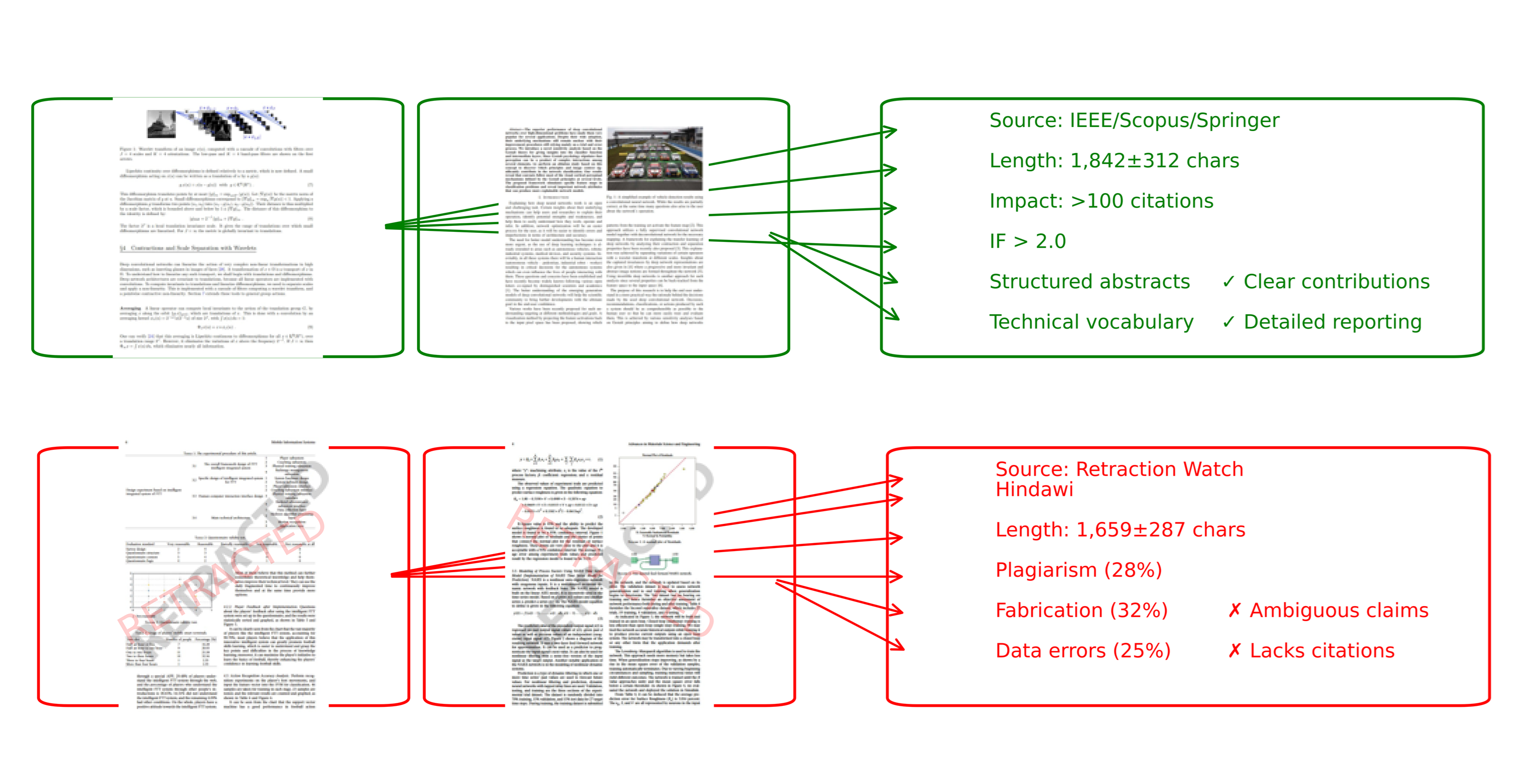}}
\vspace{-8mm}
\caption{Visual comparison between high-impact and retracted papers in our dataset. Good papers tend to be longer, from reputable sources, and display structured academic language. Retracted papers often show signs of poor writing, shorter lengths, and common retraction causes such as plagiarism and fabrication. }
\label{fig:papersamples}
\vspace{-5mm}
\end{figure}


These results have significant insights. The Neural Network with SBERT is still the best option for organizations that demand the highest accuracy at any computational cost. FastText with SVM offers a great balance in situations where processing speed is crucial. The most reliable performance across all embedding types is Random Forest, which qualifies for widespread use. Future research could look into domain-specific adaptions for various academic subjects or hybrid techniques that combine the advantages of several embeddings.

Some observations that support our findings are shown by the comparative analysis: (1) Retracted publications accurately recognized showed the citation anomalies and linguistic patterns; (2) High-impact papers correctly classified (82-91\textit{\%}  accuracy) showed expected traits of methodological rigor in their TF-IDF/SBERT embeddings. But two edge cases continue to exist: (3) Unconventional but valid methods were the main source of false negatives (2.8– 4.1\textit{\%}), especially in theoretical fields, which is a known drawback of citation-based screening; and (4) retracted papers from high-impact journals that successfully imitated strict writing conventions of false positives (1.9 – 3.3\textit{\%}).  Even though our neural network used hierarchical feature learning to reduce these error by 18 – 32\textit{\%}, these examples show that include semantic embedding techniques and metadata analysis are necessary for the best screening systems. Figure~\ref{fig:papersamples} visualizes the examples of ``good'' and ``non-good'' categories.

\section{Conclusion}
\label{sec:conclusion}

In this paper, we investigate the usage of machine learning and linguistic analysis to distinguish between retracted literature and high-quality academic papers. Using Sentence-BERT embeddings, our neural network model achieved 87.72\% classification accuracy, significantly outperforming more conventional methods like Support Vector Machine (82.94\%) and Random Forest (84.72\%). The superior performance of deep learning highlights its ability to capture subtle semantic patterns in academic writing. Complementary word cloud analysis revealed distinct linguistic signatures: retracted papers heavily relied on vague, process-oriented language like ``patient'', ``control'', and ``level'', whereas high-impact papers consistently used precise, methodology-focused terms like ``network'', ``algorithm'', and ``performance''. Together, these results demonstrate that both quantitative features and qualitative linguistic patterns provide reliable indicators of research quality, suggesting practical applications for automated screening of questionable submissions and support for peer-review workflows.

For future work, we plan to pursue the promising directions emerging from this work. First, retracted articles often show unusual citation patterns, such as abrupt spikes or anomalous co-citation networks, suggesting that integrating citation analysis could enhance detection accuracy. Second, expanding linguistic analysis to include stylistic markers such as passive voice and hedging language may further improve discriminatory power. Third, validation across diverse academic domains is also essential, given variations in disciplinary norms and retraction practices. Last but not least, practical deployment could involve developing plugins for journal submission systems or tools for preprint screening.


\bibliographystyle{splncs04}
\bibliography{sn-bibliography}

\end{document}